\documentclass{article} 
\usepackage{iclr2017_conference,times}
\usepackage{hyperref}
\usepackage{url}

\usepackage{amssymb,amsmath}
\DeclareMathOperator*{\argmin}{argmin}
\DeclareMathOperator*{\argmax}{argmax}
\usepackage{natbib}
\def\bibfont{\footnotesize}

\usepackage{graphicx}
\usepackage{caption}
\usepackage{subcaption}
\usepackage{float}

\usepackage{amsthm}
\newtheorem{theorem}{Theorem}
\newtheorem{lemma}{Lemma}

\title{Improving Sampling from Generative \newline Autoencoders with Markov
Chains}

\author{Antonia Creswell, Kai Arulkumaran \& Anil A. Bharath
\\Department of Bioengineering\\Imperial College London\\London SW7 2BP, UK\\
\texttt{\{ac2211,ka709,aab01\}@ic.ac.uk}
}

\begin{document}

\maketitle

\begin{abstract}
We focus on generative autoencoders, such as variational or adversarial
autoencoders, which jointly learn a generative model alongside an
inference model. Generative autoencoders are those which are trained to
softly enforce a prior on the latent distribution learned by the
inference model. We call the distribution to which the inference model
maps observed samples, the \emph{learned latent distribution}, which may
not be consistent with the prior. We formulate a Markov chain Monte
Carlo (MCMC) sampling process, equivalent to iteratively decoding and
encoding, which allows us to sample from the learned latent
distribution. Since, the generative model learns to map from the
\textit{learned} latent distribution, rather than the prior, we may use
MCMC to improve the quality of samples drawn from the
\textbf{generative} model, especially when the learned latent
distribution is far from the prior. Using MCMC sampling, we are able to
reveal previously unseen differences between generative autoencoders
trained either with or without a denoising criterion.
\end{abstract}

\section{Introduction}\label{introduction}

Unsupervised learning has benefited greatly from the introduction of
deep generative models. In particular, the introduction of generative
adversarial networks (GANs) \citep{goodfellow2014generative} and
variational autoencoders (VAEs)
\citep{kingma2013auto, rezende2014stochastic} has led to a plethora of
research into learning latent variable models that are capable of
generating data from complex distributions, including the space of
natural images \citep{radford2015unsupervised}. Both of these models,
and their extensions, operate by placing a prior distribution, \(P(Z)\),
over a latent space \(Z \subseteq \mathbb{R}^b\), and learn mappings
from the latent space, \(Z\), to the space of the observed data,
\(X \subseteq \mathbb{R}^a\).

We are interested in autoencoding generative models, models which learn
not just the generative mapping \(Z \mapsto X\), but also the
inferential mapping \(X \mapsto Z\). Specifically, we define
\emph{generative autoencoders} as autoencoders which softly constrain
their latent distribution, to match a specified prior distribution,
\(P(Z)\). This is achieved by minimising a loss,
\(\mathcal{L}_{prior}\), between the latent distribution and the prior.
This includes VAEs \citep{kingma2013auto, rezende2014stochastic},
extensions of VAEs \citep{kingma2016improving}, and also adversarial
autoencoders (AAEs) \citep{makhzani2015adversarial}. Whilst other
autoencoders also learn an encoding function,
\(e: \mathbb{R}^a \rightarrow Z\), together with a decoding function,
\(d: \mathbb{R}^b \rightarrow X\), the latent space is not necessarily
constrained to conform to a specified probability distribution. This is
the key distinction for generative autoencoders; both \(e\) and \(d\)
can still be deterministic functions \citep{makhzani2015adversarial}.

The functions \(e\) and \(d\) are defined for any input from
\(\mathbb{R}^a\) and \(\mathbb{R}^b\) respectively, however the outputs
of the functions may be constrained practically by the type of functions
that \(e\) and \(d\) are, such that \(e\) maps to
\(Z \subseteq \mathbb{R}^b\) and \(d\) maps to
\(X \subseteq \mathbb{R}^a\). During training however, the encoder,
\(e\) is only fed with training data samples, \(\mathbf{x} \in X\) and
the decoder, \(d\) is only fed with samples from the encoder,
\(\mathbf{z} \in Z\), and so the encoder and decoder learn mappings
between \(X\) and \(Z\).

The process of encoding and decoding may be interpreted as sampling the
conditional probabilities \(Q_{\phi}(Z|X)\) and \(P_{\theta}(X|Z)\)
respectively. The conditional distributions may be sampled using the
encoding and decoding functions \(e(X;\phi)\) and \(d(Z;\theta)\), where
\(\phi\) and \(\theta\) are learned parameters of the encoding and
decoding functions respectively. The decoder of a generative autoencoder
may be used to generate new samples that are consistent with the data.
There are two traditional approaches for sampling generative
autoencoders:

\textbf{Approach 1} \citep{bengio2014deep}:

\[\mathbf{x}_0 \sim P(X), \hspace{5mm} \mathbf{z}_0 \sim Q_{\phi}(Z|X=\mathbf{x}_0), \hspace{5mm} \mathbf{x}_1 \sim P_{\theta}(X|Z=\mathbf{z}_0)\]

where \(P(X)\) is the data generating distribution. However, this
approach is likely to generate samples similar to those in the training
data, rather than generating novel samples that are consistent with the
training data.

\textbf{Approach 2}
\citep{kingma2013auto, makhzani2015adversarial, rezende2014stochastic}:
\[ \mathbf{z_0} \sim P(Z), \hspace{5mm} \mathbf{x_0} \sim P_{\theta}(X|Z=\mathbf{z}_0)\]

where \(P(Z)\) is the prior distribution enforced during training and
\(P_{\theta}(X|Z)\) is the decoder trained to map samples drawn from
\(Q_{\phi}(Z|X)\) to samples consistent with \(P(X)\). This approach
assumes that \(\int Q_{\phi}(Z|X)P(X) dX = P(Z)\), suggesting that the
encoder maps all data samples from \(P(X)\) to a distribution that
matches the prior distribution, \(P(Z)\). However, it is not always true
that \(\int Q_{\phi}(Z|X)P(X)dX = P(Z)\). Rather \(Q_{\phi}(Z|X)\) maps
data samples to a distribution which we call, \(\hat{P}(Z)\):

\[ \int Q_{\phi}(Z|X)P(X)dX = \hat{P}(Z) \]

where it is not necessarily true that \(\hat{P}(Z)=P(Z)\) because the
prior is only softly enforced. The decoder, on the other hand, is
trained to map encoded data samples (i.e.~samples from
\(\int Q_{\phi}(Z|X)P(X) dX\)) to samples from \(X\) which have the
distribution \(P(X)\). If the encoder maps observed samples to latent
samples with the distribution \(\hat{P}(Z)\), rather than the desired
prior distribution, \(P(Z)\), then:

\[\int P_{\theta}(X|Z)P(Z) dZ \neq P(X)\]

This suggests that samples drawn from the decoder, \(P_{\theta}(X|Z)\),
conditioned on samples drawn from the prior, \(P(Z)\), may not be
consistent with the data generating distribution, \(P(X)\). However, by
conditioning on \(\hat{P}(Z)\):

\[\int P_{\theta}(X|Z)\hat{P}(Z) dZ = P(X)\]

This suggests that to obtain more realistic generations, latent samples
should be drawn via \(\mathbf{z} \sim \hat{P}(Z)\) rather than
\(\mathbf{z} \sim P(Z)\), followed by
\(\mathbf{x} \sim P_{\theta}(X|Z)\). A limited number of latent samples
may be drawn from \(\hat{P}(Z)\) using the first two steps in Approach 1
- however this has the drawbacks discussed in Approach 1. We introduce
an alternative method for sampling from \(\hat{P}(Z)\) which does not
have the same drawbacks.

\begin{figure}
\centering
\includegraphics[width=0.55000\textwidth]{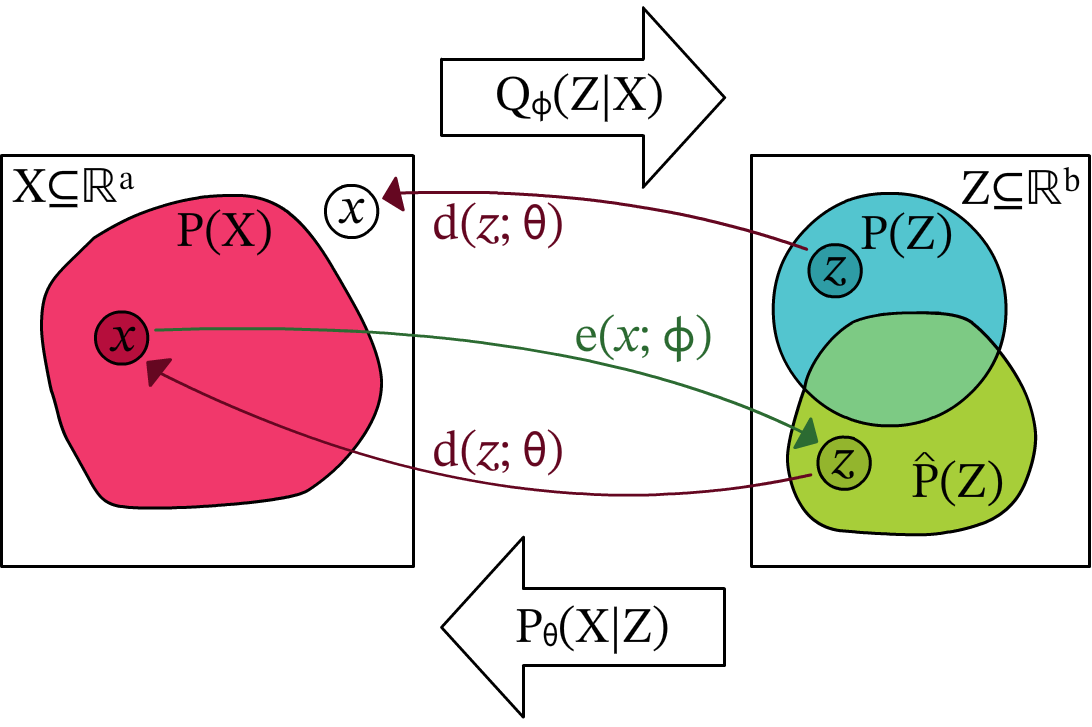}
\caption{\(P(X)\) is the data generating distribution. We may access
some samples from \(P(X)\) by drawing samples from the training data.
\(Q_{\phi}(Z|X)\) is the conditional distribution, modeled by an
encoder, which maps samples from \(\mathbb{R}^a\) to samples in
\(\mathbb{R}^b\). An ideal encoder maps samples from \(P(X)\) to a
known, prior distribution \(P(Z)\): in reality the encoder maps samples
from \(P(X)\) to an unknown distribution \(\hat{P}(Z)\).
\(P_{\theta}(X|Z)\) is a conditional distribution, modeled by a decoder,
which maps samples from \(\mathbb{R}^b\) to \(\mathbb{R}^a\). During
training the decoder learns to map samples drawn from \(\hat{P}(Z)\) to
\(P(X)\) rather than samples drawn from \(P(Z)\) because the decoder
only sees samples from \(\hat{P}(Z)\). Regularisation on the latent
space only encourages \(\hat{P}(Z)\) to be close to \(P(Z)\). Note that
if \(\mathcal{L}_{prior}\) is optimal, then \(\hat{P}(Z)\) overlaps
fully with \(P(Z)\).\label{ideal}}
\end{figure}

\begin{figure}
  \centering
  \begin{subfigure}[]{0.97\linewidth}
    \includegraphics[width=\linewidth]{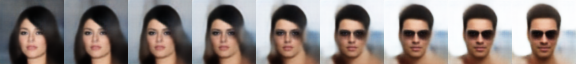}
    \subcaption{VAE (initial)}
  \end{subfigure}
  \\[0.4ex]
  \begin{subfigure}[]{0.97\linewidth}
    \includegraphics[width=\linewidth]{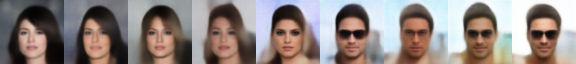}
    \subcaption{VAE (5 steps)}
  \end{subfigure}
  \\[2ex]
  \begin{subfigure}[]{0.97\linewidth}
    \includegraphics[width=\linewidth]{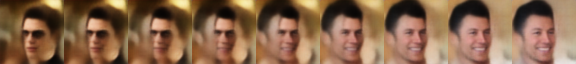}
    \subcaption{VAE (initial)}
  \end{subfigure}
  \\[0.4ex]
  \begin{subfigure}[]{0.97\linewidth}
    \includegraphics[width=\linewidth]{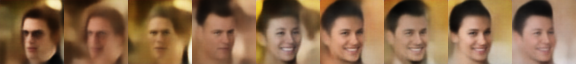}
    \subcaption{VAE (5 steps)}
  \end{subfigure}
  \caption{\textbf{Prior work:} Spherically interpolating \citep{white2016sampling} between two faces using a VAE (a, c). In (a), the attempt to gradually generate sunglasses results in visual artifacts around the eyes. In (c), the model fails to properly capture the desired change in orientation of the face, resulting in three partial faces in the middle of the interpolation. \textbf{This work:} (b) and (d) are the result of 5 steps of MCMC sampling applied to the latent samples that were used to generate the original interpolations, (a) and (c). In (b), the discolouration around the eyes disappears, with the model settling on either generating or not generating glasses. In (d), the model moves away from multiple faces in the interpolation by producing new faces with appropriate orientations.}
  \label{vae_interp}
\end{figure}

Our main contribution is the formulation of a Markov chain Monte Carlo
(MCMC) sampling process for generative autoencoders, which allows us to
sample from \(\hat{P}(Z)\). By iteratively sampling the chain, starting
from an arbitrary \(\mathbf{z}_{t=0} \in \mathbb{R}^b\), the chain
converges to \(\mathbf{z}_{t \rightarrow \infty} \sim \hat{P}(Z)\),
allowing us to draw latent samples from \(\hat{P}(Z)\) after several
steps of MCMC sampling. From a practical perspective, this is achieved
by iteratively decoding and encoding, which may be easily applied to
existing generative autoencoders. Because \(\hat{P}(Z)\) is optimised to
be close to \(P(Z)\), the initial sample, \(\mathbf{z}_{t=0}\) can be
drawn from \(P(Z)\), improving the quality of the samples within a few
iterations.

When interpolating between latent encodings, there is no guarantee that
\(\mathbf{z}\) stays within high density regions of \(\hat{P}(Z)\).
Previously, this has been addressed by using spherical, rather than
linear interpolation of the high dimensional \(Z\) space
\citep{white2016sampling}. However, this approach attempts to keep
\(\mathbf{z}\) within \(P(Z)\), rather than trying to sample from
\(\hat{P}(Z)\). By instead applying several steps of MCMC sampling to
the interpolated \(\mathbf{z}\) samples before sampling
\(P_{\theta}(X|Z)\), unrealistic artifacts can be reduced (see Figure
\ref{vae_interp}). Whilst most methods that aim to generate realistic
samples from \(X\) rely on adjusting encodings of the observed data
\citep{white2016sampling}, our use of MCMC allows us to walk any latent
sample to more probable regions of the learned latent distribution,
resulting in more convincing generations. We demonstrate that the use of
MCMC sampling improves generations from both VAEs and AAEs with
high-dimensional \(Z\); this is important as previous studies have shown
that the dimensionality of \(Z\) should be scaled with the intrinsic
latent dimensionality of the observed data.

Our second contribution is the modification of the proposed transition
operator for the MCMC sampling process to denoising generative
autoencoders. These are generative autoencoders trained using a
denoising criterion, \citep{seung1997learning, vincent2008extracting}.
We reformulate our original MCMC sampling process to incorporate the
noising and denoising processes, allowing us to use MCMC sampling on
denoising generative autoencoders. We apply this sampling technique to
two models. The first is the denoising VAE (DVAE) introduced by Im et
al. \citeyearpar{im2015denoising}. We found that MCMC sampling revealed
benefits of the denoising criterion. The second model is a denoising AAE
(DAAE), constructed by applying the denoising criterion to the AAE.
There were no modifications to the cost function. For both the DVAE and
the DAAE, the effects of the denoising crtierion were not immediately
obvious from the initial samples. Training generative autoencoders with
a denoising criterion reduced visual artefacts found both in generations
and in interpolations. The effect of the denoising criterion was
revealed when sampling the denoising models using MCMC sampling.

\section{Background}\label{background}

One of the main tasks in machine learning is to learn explanatory
factors for observed data, commonly known as inference. That is, given a
data sample \(\mathbf{x} \in X \subseteq \mathbb{R}^a\), we would like
to find a corresponding latent encoding
\(\mathbf{z} \in Z \subseteq \mathbb{R}^b\). Another task is to learn
the inverse, generative mapping from a given \(\mathbf{z}\) to a
corresponding \(\mathbf{x}\). In general, coming up with a suitable
criterion for learning these mappings is difficult. Autoencoders solve
both tasks efficiently by jointly learning an inferential mapping
\(e(X;\phi)\) and generative mapping \(d(Z;\theta)\), using unlabelled
data from \(X\) in a self-supervised fashion \citep{kingma2013auto}. The
basic objective of all autoencoders is to minimise a reconstruction
cost, \(\mathcal{L}_{reconstruct}\), between the original data, \(X\),
and its reconstruction, \(d(e(X;\phi);\theta)\). Examples of
\(\mathcal{L}_{reconstruct}\) include the squared error loss,
\(\frac{1}{2}\sum_{n=1}^N\Vert d(e(\mathbf{x}_n;\phi);\theta) - \mathbf{x}_n\Vert^2\),
and the cross-entropy loss,
\(\mathcal{H}[P(X)\Vert P(d(e(X;\phi);\theta))] = -\sum_{n=1}^N\mathbf{x}_n\log(d(e(\mathbf{x}_n; \phi);\theta)) + (1 - \mathbf{x}_n)\log(1 - d(e(\mathbf{x}_n;\phi);\theta))\).

Autoencoders may be cast into a probablistic framework, by considering
samples \(\mathbf{x} \sim P(X)\) and \(\mathbf{z} \sim P(Z)\), and
attempting to learn the conditional distributions \(Q_{\phi}(Z|X)\) and
\(P_{\theta}(X|Z)\) as \(e(X;\phi)\) and \(d(Z;\theta)\) respectively,
with \(\mathcal{L}_{reconstruct}\) representing the negative
log-likelihood of the reconstruction given the encoding
\citep{bengio2009learning}. With any autoencoder, it is possible to
create novel \(\mathbf{x} \in X\) by passing a \(\mathbf{z} \in Z\)
through \(d(Z;\theta)\), but we have no knowledge of appropriate choices
of \(\mathbf{z}\) beyond those obtained via \(e(X;\phi)\). One solution
is to constrain the latent space to which the encoding model maps
observed samples. This can be achieved by an additional loss,
\(\mathcal{L}_{prior}\), that penalises encodings far away from a
specified prior distribution, \(P(Z)\). We now review two types of
generative autoencoders, VAEs
\citep{kingma2013auto, rezende2014stochastic} and AAEs
\citep{makhzani2015adversarial}, which each take different approaches to
formulating \(\mathcal{L}_{prior}\).

\subsection{Generative autoencoders}\label{generative-autoencoders}

Consider the case where \(e\) is constructed with stochastic neurons
that can produce outputs from a specified probability distribution, and
\(\mathcal{L}_{prior}\) is used to constrain the distribution of outputs
to \(P(Z)\). This leaves the problem of estimating the gradient of the
autoencoder over the expectation \(\mathbb{E}_{Q_{\phi}(Z|X)}\), which
would typically be addressed with a Monte Carlo method. VAEs sidestep
this by constructing latent samples using a deterministic function and a
source of noise, moving the source of stochasticity to an input, and
leaving the network itself deterministic for standard gradient
calculations---a technique commonly known as the reparameterisation
trick \citep{kingma2013auto}. \(e(X;\phi)\) then consists of a
deterministic function, \(e_{rep}(X;\phi)\), that outputs parameters for
a probability distribution, plus a source of noise. In the case where
\(P(Z)\) is a diagonal covariance Gaussian, \(e_{rep}(X;\phi)\) maps
\(\mathbf{x}\) to a vector of means,
\(\boldsymbol{\mu} \in \mathbb{R}^b\), and a vector of standard
deviations, \(\boldsymbol{\sigma} \in \mathbb{R}_+^b\), with the noise
\(\boldsymbol{\epsilon} \sim \mathcal{N}(\mathbf{0}, \mathbf{I})\). Put
together, the encoder outputs samples
\(\mathbf{z} = \boldsymbol{\mu} + \boldsymbol{\epsilon}\odot\boldsymbol{\sigma}\),
where \(\odot\) is the Hadamard product. VAEs attempt to make these
samples from the encoder match up with \(P(Z)\) by using the KL
divergence between the parameters for a probability distribution
outputted by \(e_{rep}(X;\phi)\), and the parameters for the prior
distribution, giving
\(\mathcal{L}_{prior} = D_{KL}[Q_{\phi}(Z|X)\Vert P(Z)]\). A
multivariate Gaussian has an analytical KL divergence that can be
further simplified when considering the unit Gaussian, resulting in
\(\mathcal{L}_{prior} = \frac{1}{2} \sum_{n=1}^N \boldsymbol{\mu}^2 + \boldsymbol{\sigma}^2 - \log(\boldsymbol{\sigma}^2) - \mathbf{1}\).

Another approach is to deterministically output the encodings
\(\mathbf{z}\). Rather than minimising a metric between probability
distributions using their parameters, we can turn this into a density
ratio estimation problem where the goal is to learn a conditional
distribution, \(Q_{\phi}(Z|X)\), such that the distribution of the
encoded data samples, \(\hat{P}(Z)=\int Q_{\phi}(Z|X)P(X) dX\), matches
the prior distribution, \(P(Z)\). The GAN framework solves this density
ratio estimation problem by transforming it into a class estimation
problem using two networks \citep{goodfellow2014generative}. The first
network in GAN training is the discriminator network, \(D_{\psi}\),
which is trained to maximise the log probability of samples from the
``real'' distribution, \(\mathbf{z} \sim P(Z)\), and minimise the log
probability of samples from the ``fake'' distribution,
\(\mathbf{z} \sim Q_{\phi}(Z|X)\). In our case \(e(X;\phi)\) plays the
role of the second network, the generator network, \(G_{\phi}\), which
generates the ``fake'' samples.\footnote{We adapt the variables to
  better fit the conventions used in the context of autoencoders.} The
two networks compete in a minimax game, where \(G_{\phi}\) receives
gradients from \(D_{\psi}\) such that it learns to better fool
\(D_{\psi}\). The training objective for both networks is given by
\(\mathcal{L}_{prior} = \argmin_{\phi} \argmax_{\psi} \mathbb{E}_{P(Z)}[\log(D_{\psi}(Z))] + \mathbb{E}_{P(X)}[\log(1 - D_{\psi}(G_{\phi}(X)))]=\argmin_{\phi} \argmax_{\psi} \mathbb{E}_{P(Z)}[\log(D_{\psi}(Z))]+\mathbb{E}_{Q_{\phi}(Z|X)P(X)}\log[1-D_{\psi}(Z)]\).
This formulation can create problems during training, so instead
\(G_{\phi}\) is trained to minimise \(-\log(D_{\psi}(G_{\phi}(X)))\),
which provides the same fixed point of the dynamics of \(G_{\phi}\) and
\(D_{\psi}\). The result of applying the GAN framework to the encoder of
an autoencoder is the deterministic AAE \citep{makhzani2015adversarial}.

\subsection{Denoising autoencoders}\label{denoising-autoencoders}

\begin{figure}
\centering
\includegraphics[width=0.66000\textwidth]{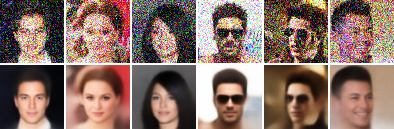}
\caption{Reconstructions of faces from a DVAE trained with additive
Gaussian noise: \(Q(\tilde{X}|X) = \mathcal{N}(X, 0.25\mathbf{I})\). The
model successfully recovers much of the detail from the noise-corrupted
images.\label{reconstructions}}
\end{figure}

In a more general viewpoint, generative autoencoders fulfill the purpose
of learning useful representations of the observed data. Another widely
used class of autoencoders that achieve this are denoising autoencoders
(DAEs), which are motivated by the idea that learned features should be
robust to ``partial destruction of the input''
\citep{vincent2008extracting}. Not only does this require encoding the
inputs, but capturing the statistical dependencies between the inputs so
that corrupted data can be recovered (see Figure \ref{reconstructions}).
DAEs are presented with a corrupted version of the input,
\(\mathbf{\tilde{x}} \in \tilde{X}\), but must still reconstruct the
original input, \(\mathbf{x} \in X\), where the noisy inputs are created
through sampling \(\mathbf{\tilde{x}} \sim C(\tilde{X}|X)\), a
corruption process. The denoising criterion, \(\mathcal{L}_{denoise}\),
can be applied to any type of autoencoder by replacing the
straightforward reconstruction criterion,
\(\mathcal{L}_{reconstruct}(X, d(e(X;\phi);\theta))\), with the
reconstruction criterion applied to noisy inputs:
\(\mathcal{L}_{reconstruct}(X, d(e(\tilde{X};\phi);\theta))\). The
encoder is now used to model samples drawn from
\(Q_{\phi}(Z|\tilde{X})\). As such, we can construct \emph{denoising
generative autoencoders} by training autoencoders to minimise
\(\mathcal{L}_{denoise} + \mathcal{L}_{prior}\).

One might expect to see differences in samples drawn from denoising
generative autoencoders and their non-denoising counterparts. However,
Figures \ref{all_samples} and \ref{interp_aae} show that this is not the
case. Im et al. \citeyearpar{im2015denoising} address the case of DVAEs,
claiming that the noise mapping requires adjusting the original VAE
objective function. Our work is orthogonal to theirs, and others which
adjust the training or model \citep{kingma2016improving}, as we focus
purely on sampling from generative autoencoders after training. We claim
that the existing practice of drawing samples from generative
autoencoders conditioned on \(\mathbf{z} \sim P(Z)\) is suboptimal, and
the quality of samples can be improved by instead conditioning on
\(\mathbf{z} \sim \hat{P}(Z)\) via MCMC sampling.

\section{Markov sampling}\label{markov-sampling}

We now consider the case of sampling from generative autoencoders, where
\(d(Z;\theta)\) is used to draw samples from \(P_{\theta}(X|Z)\). In
Section \ref{introduction}, we showed that it was important, when
sampling \(P_{\theta}(X|Z)\), to condition on \(\mathbf{z}\)'s drawn
from \(\hat{P}(Z)\), rather than \(P(Z)\) as is often done in practice.
However, we now show that for any initial
\(\mathbf{z}_0 \in Z_0 = \mathbb{R}^b\), Markov sampling can be used to
produce a chain of samples \(\mathbf{z}_t\), such that as
\(t \rightarrow \infty\), produces samples \(\mathbf{z}_t\) that are
from the distribution \(\hat{P}(Z)\), which may be used to draw
meaningful samples from \(P_{\theta}(X|Z)\), conditioned on
\(\mathbf{z} \sim \hat{P}(Z)\). To speed up convergence we can
initialise \(\mathbf{z}_0\) from a distribution close to \(\hat{P}(Z)\),
by drawing \(\mathbf{z}_0 \sim P(Z)\).

\subsection{Markov sampling process}\label{markov-sampling-process}

A generative autoencoder can be sampled by the following process:
\[\mathbf{z}_0 \in Z_0 = \mathbb{R}^b, \hspace{5mm} \mathbf{x}_{t+1} \sim P_{\theta}(X|Z_t), \hspace{5mm} \mathbf{z}_{t+1} \sim Q_{\phi}(Z|X_{t+1})\]
This allows us to define a Markov chain with the transition operator

\begin{equation}
\label{trans_operator}
T(Z_{t+1}|Z_t) = \int Q_{\phi}(Z_{t+1}|X)P_{\theta}(X|Z_t) dX
\end{equation}

for \(t \geq 0\).

Drawing samples according to the transition operator \(T(Z_{t+1}|Z_t)\)
produces a Markov chain. For the transition operator to be homogeneous,
the parameters of the encoding and decoding functions are fixed during
sampling.

\subsection{Convergence properties}\label{convergence-properties}

We now show that the stationary distribution of sampling from the Markov
chain is \(\hat{P}(Z)\).

\begin{theorem}
\label{convergence}
If $T(Z_{t+1}|Z_t)$ defines an ergodic Markov chain, $\{Z_1, Z_2 ... Z_t\}$, then the chain will converge to a stationary distribution, $\Pi(Z)$, from any arbitrary initial distribution. The stationary distribution $\Pi(Z) = \hat{P}(Z)$.
\end{theorem}

The proof of Theorem \ref{convergence} can be found in
\citep{rosenthal2001review}.

\begin{lemma}
\label{ergodic}
$T(Z_{t+1}|Z_t)$ defines an ergodic Markov chain.
\end{lemma}

\begin{proof}
For a Markov chain to be ergodic it must be both irreducible (it is possible to get from any state to any other state in a finite number of steps) and aperiodic (it is possible to get from any state to any other state without having to pass through a cycle). To satisfy these requirements, it is more than sufficient to show that $T(Z_{t+1}|Z_t) > 0$, since every $\mathbf{z} \in Z$ would be reachable from every other $\mathbf{z} \in Z$. We show that $P_{\theta}(X|Z) > 0$ and $Q_{\phi}(Z|X) > 0$, giving $T(Z_{t+1}|Z_t) > 0$, providing the proof of this in Section \ref{proof} of the supplementary material.
\end{proof}

\begin{lemma}
\label{stationary}
The stationary distribution of the chain defined by $T(Z_{t+1}|Z_t)$ is $\Pi(Z) = \hat{P}(Z)$.
\end{lemma}

\begin{proof}
For the transition operator defined in Equation (\ref{trans_operator}), the asymptotic distribution to which $T(Z_{t+1}|Z_t)$ converges to is $\hat{P}(Z)$, because $\hat{P}(Z)$ is, by definition, the marginal of the joint distribution $Q_{\phi}(Z|X)P(X)$, over which the $\mathcal{L}_{prior}$ used to learn the conditional distribution $Q_{\phi}(Z|X)$.\end{proof}

Using Lemmas \ref{ergodic} and \ref{stationary} with Theorem
\ref{convergence}, we can say that the Markov chain defined by the
transition operator in Equation (\ref{trans_operator}) will produce a
Markov chain that converges to the stationary distribution
\(\Pi(Z) = \hat{P}(Z)\).

\subsection{Extension to denoising generative
autoencoders}\label{extension-to-denoising-generative-autoencoders}

A denoising generative autoencoder can be sampled by the following
process:
\[\mathbf{z}_0 \in Z_0 = \mathbb{R}^b, \hspace{5mm} \mathbf{x}_{t+1} \sim P_{\theta}(X|Z_t), \hspace{5mm} \mathbf{\tilde{x}}_{t+1} \sim C(\tilde{X}|X_{t+1}), \hspace{5mm}\mathbf{z}_{t+1} \sim Q_{\phi}(Z|\tilde{X}_{t+1}).\]
This allows us to define a Markov chain with the transition operator

\begin{equation}
\label{denoising_trans_operator}
T(Z_{t+1}|Z_t) = \int Q_{\phi}(Z_{t+1}|\tilde{X})C(\tilde{X}|X)P_{\theta}(X|Z_t) dXd\tilde{X}
\end{equation}

for \(t \geq 0\).

The same arguments for the proof of convergence of Equation
(\ref{trans_operator}) can be applied to Equation
(\ref{denoising_trans_operator}).

\subsection{Related work}\label{related-work}

Our work is inspired by that of Bengio et al.
\citeyearpar{bengio2013generalized}; denoising autoencoders are cast
into a probabilistic framework, where \(P_{\theta}(X|\tilde{X})\) is the
\textit{denoising} (decoder) distribution and \(C(\tilde{X}|X)\) is the
\textit{corruption} (encoding) distribution. \(\tilde{X}\) represents
the space of corrupted samples. Bengio et al.
\citeyearpar{bengio2013generalized} define a transition operator of a
Markov chain -- using these conditional distributions -- whose
stationary distribution is \(P(X)\) under the assumption that
\(P_{\theta}(X|\tilde{X})\) perfectly denoises samples. The chain is
initialised with samples from the training data, and used to generate a
chain of samples from \(P(X)\). This work was generalised to include a
corruption process that mapped data samples to latent variables
\citep{bengio2014deep}, to create a new type of network called
Generative Stochastic Networks (GSNs). However in GSNs
\citep{bengio2014deep} the latent space is not regularised with a prior.

Our work is similar to several approaches proposed by Bengio et al.
\citetext{\citeyear{bengio2013generalized}; \citeyear{bengio2014deep}}
and Rezende et al. \citep{rezende2014stochastic}. Both Bengio et al. and
Rezende et al. define a transition operator in terms of \(X_t\) and
\(X_{t-1}\). Bengio et al. generate samples with an initial \(X_0\)
drawn from the observed data, while Rezende et al. reconstruct samples
from an \(X_0\) which is a corrupted version of a data sample. In
contrasts to Bengio et al. and Rezende et al., in this work we define
the transition operator in terms of \(Z_{t+1}\) and \(Z_t\), initialise
samples with a \(Z_0\) that is drawn from a prior distribution we can
directly sample from, and then sample \(X_1\) conditioned on \(Z_0\).
Although the initial samples may be poor, we are likely to generate a
novel \(X_1\) on the first step of MCMC sampling, which would not be
achieved using Bengio et al.'s or Rezende et al.'s approach. We are able
draw initial \(Z_0\) from a prior because we constrain \(\hat{P}(Z)\) to
be close to a prior distribution \(P(Z)\); in Bengio et al. a latent
space is either not explicitly modeled \citep{bengio2013generalized} or
it is not constrained \citep{bengio2014deep}.

Further, Rezende et al. \citeyearpar{rezende2014stochastic} explicitly
assume that the distribution of latent samples drawn from
\(Q_{\phi}(Z|X)\) matches the prior, \(P(Z)\). Instead, we assume that
samples drawn from \(Q_{\phi}(Z|X)\) have a distribution \(\hat{P}(Z)\)
that does not necessarily match the prior, \(P(Z)\). We propose an
alternative method for sampling \(\hat{P}(Z)\) in order to improve the
quality of generated image samples. Our motivation is also different to
Rezende et al. \citeyearpar{rezende2014stochastic} since we use sampling
to generate improved, novel data samples, while they use sampling to
denoise corrupted samples.

\subsection{Effect of regularisation
method}\label{effect-of-regularisation-method}

The choice of \(\mathcal{L}_{prior}\) may effect how much improvement
can be gained when using MCMC sampling, assuming that the optimisation
process converges to a reasonable solution. We first consider the case
of VAEs, which minimise \(D_{KL}[Q_{\phi}(Z|X)\Vert P(Z)]\). Minimising
this KL divergence penalises the model \(\hat{P}(Z)\) if it contains
samples that are outside the support of the true distribution \(P(Z)\),
which might mean that \(\hat{P}(Z)\) captures only a part of \(P(Z)\).
This means that when sampling \(P(Z)\), we may draw from a region that
is not captured by \(\hat{P}(Z)\). This suggests that MCMC sampling can
improve samples from trained VAEs by walking them towards denser regions
in \(\hat{P}(Z)\).

Generally speaking, using the reverse KL divergence during training,
\(D_{KL}[P(Z)\Vert Q_{\phi}(Z|X)]\), penalises the model
\(Q_{\phi}(Z|X)\) if \(P(Z)\) produces samples that are outside of the
support of \(\hat{P}(Z)\). By minimising this KL divergence, most
samples in \(P(Z)\) will likely be in \(\hat{P}(Z)\) as well. AAEs, on
the other hand are regularised using the JS entropy, given by
\(\frac{1}{2}D_{KL}[P(Z)\Vert\frac{1}{2}(P(Z) + Q_{\phi}(Z|X))] + \frac{1}{2}D_{KL}[Q_{\phi}(Z|X)\Vert\frac{1}{2}(P(Z) + Q_{\phi}(Z|X))]\).
Minimising this cost function attempts to find a compromise between the
aforementioned extremes. However, this still suggests that some samples
from \(P(Z)\) may lie outside \(\hat{P}(Z)\), and so we expect AAEs to
also benefit from MCMC sampling.

\section{Experiments}\label{experiments}

\subsection{Models}\label{models}

We utilise the deep convolutional GAN (DCGAN)
\citep{radford2015unsupervised} as a basis for our autoencoder models.
Although the recommendations from Radford et al.
\citeyearpar{radford2015unsupervised} are for standard GAN
architectures, we adopt them as sensible defaults for an autoencoder,
with our encoder mimicking the DCGAN's discriminator, and our decoder
mimicking the generator. The encoder uses strided convolutions rather
than max-pooling, and the decoder uses fractionally-strided convolutions
rather than a fixed upsampling. Each convolutional layer is succeeded by
spatial batch normalisation \citep{ioffe2015batch} and ReLU
nonlinearities, except for the top of the decoder which utilises a
sigmoid function to constrain the output values between 0 and 1. We
minimise the cross-entropy between the original and reconstructed
images. Although this results in blurry images in regions which are
ambiguous, such as hair detail, we opt not to use extra loss functions
that improve the visual quality of generations
\citep{larsen2015autoencoding, dosovitskiy2016generating, lamb2016discriminative}
to avoid confounding our results.

Although the AAE is capable of approximating complex probabilistic
posteriors \citep{makhzani2015adversarial}, we construct ours to output
a deterministic \(Q_{\phi}(Z|X)\). As such, the final layer of the
encoder part of our AAEs is a convolutional layer that deterministically
outputs a latent sample, \(\mathbf{z}\). The adversary is a
fully-connected network with dropout and leaky ReLU nonlinearities.
\(e_{rep}(X;\phi)\) of our VAEs have an output of twice the size, which
corresponds to the means, \(\boldsymbol{\mu}\), and standard deviations,
\(\boldsymbol{\sigma}\), of a diagonal covariance Gaussian distribution.
For all models our prior, \(P(Z)\), is a 200D isotropic Gaussian with
zero mean and unit variance: \(\mathcal{N}(\mathbf{0}, \mathbf{I})\).

\subsection{Datasets}\label{datasets}

Our primary dataset is the (aligned and cropped) CelebA dataset, which
consists of 200,000 images of celebrities \citep{liu2015deep}. The DCGAN
\citep{radford2015unsupervised} was the first generative neural network
model to show convincing novel samples from this dataset, and it has
been used ever since as a qualitative benchmark due to the amount and
quality of samples. In Figures \ref{svhn_samples} and
\ref{svhn_interp_vae} of the supplementary material, we also include
results on the SVHN dataset, which consists of 100,000 images of house
numbers extracted from Google Street view images
\citep{netzer2011reading}.

\subsection{Training \& evaluation}\label{training-evaluation}

For all datasets we perform the same preprocessing: cropping the centre
to create a square image, then resizing to \(64 \times 64\)px. We train
our generative autoencoders for 20 epochs on the training split of the
datasets, using Adam \citep{kingma2014adam} with \(\alpha = 0.0002\),
\(\beta_1 = 0.5\) and \(\beta_2 = 0.999\). The denoising generative
autoencoders use the additive Gaussian noise mapping
\(C(\tilde{X}|X) = \mathcal{N}(X, 0.25\mathbf{I})\). All of our
experiments were run using the Torch library
\citep{collobert2011torch7}.\footnote{Example code is available at
  \url{https://github.com/Kaixhin/Autoencoders}.}

\begin{figure}[H]
  \begin{subfigure}[]{0.21\linewidth}
    \includegraphics[width=\linewidth]{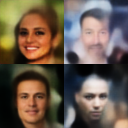}
    \subcaption{VAE (initial)}
  \end{subfigure}
  \hfill
  \begin{subfigure}[]{0.21\linewidth}
    \includegraphics[width=\linewidth]{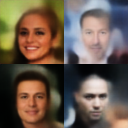}
    \subcaption{VAE (1 step)}
  \end{subfigure}
  \hfill
  \begin{subfigure}[]{0.21\linewidth}
    \includegraphics[width=\linewidth]{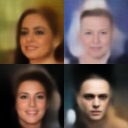}
    \subcaption{VAE (5 steps)}
  \end{subfigure}
  \hfill
  \begin{subfigure}[]{0.21\linewidth}
    \includegraphics[width=\linewidth]{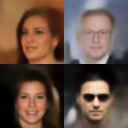}
    \subcaption{VAE (10 steps)}
  \end{subfigure}
  \\[2ex]
  \begin{subfigure}[]{0.21\linewidth}
    \includegraphics[width=\linewidth]{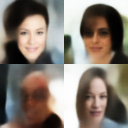}
    \subcaption{DVAE (initial)}
  \end{subfigure}
  \hfill
  \begin{subfigure}[]{0.21\linewidth}
    \includegraphics[width=\linewidth]{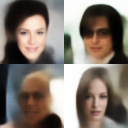}
    \subcaption{DVAE (1 step)}
  \end{subfigure}
  \hfill
  \begin{subfigure}[]{0.21\linewidth}
    \includegraphics[width=\linewidth]{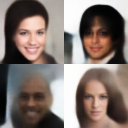}
    \subcaption{DVAE (5 steps)}
  \end{subfigure}
  \hfill
  \begin{subfigure}[]{0.21\linewidth}
    \includegraphics[width=\linewidth]{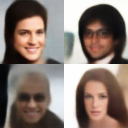}
    \subcaption{DVAE (10 steps)}
  \end{subfigure}
  \\[2ex]
  \begin{subfigure}[]{0.21\linewidth}
    \includegraphics[width=\linewidth]{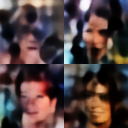}
    \subcaption{AAE (initial)}
  \end{subfigure}
  \hfill
  \begin{subfigure}[]{0.21\linewidth}
    \includegraphics[width=\linewidth]{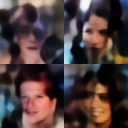}
    \subcaption{AAE (1 step)}
  \end{subfigure}
  \hfill
  \begin{subfigure}[]{0.21\linewidth}
    \includegraphics[width=\linewidth]{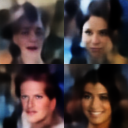}
    \subcaption{AAE (5 steps)}
  \end{subfigure}
  \hfill
  \begin{subfigure}[]{0.21\linewidth}
    \includegraphics[width=\linewidth]{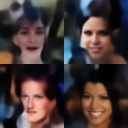}
    \subcaption{AAE (10 steps)}
  \end{subfigure}
  \\[2ex]
  \begin{subfigure}[]{0.21\linewidth}
    \includegraphics[width=\linewidth]{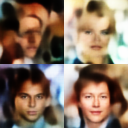}
    \subcaption{DAAE (initial)}
  \end{subfigure}
  \hfill
  \begin{subfigure}[]{0.21\linewidth}
    \includegraphics[width=\linewidth]{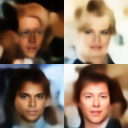}
    \subcaption{DAAE (1 step)}
  \end{subfigure}
  \hfill
  \begin{subfigure}[]{0.21\linewidth}
    \includegraphics[width=\linewidth]{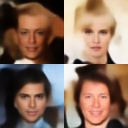}
    \subcaption{DAAE (5 steps)}
  \end{subfigure}
  \hfill
  \begin{subfigure}[]{0.21\linewidth}
    \includegraphics[width=\linewidth]{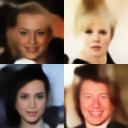}
    \subcaption{DAAE (10 steps)}
  \end{subfigure}
  \caption{Samples from a VAE (a-d), DVAE (e-h), AAE (i-l) and DAAE (m-p) trained on the CelebA dataset. (a), (e), (i) and (m) show initial samples conditioned on $\mathbf{z} \sim P(Z)$, which mainly result in recognisable faces emerging from noisy backgrounds. After 1 step of MCMC sampling, the more unrealistic generations change noticeably, and continue to do so with further steps. On the other hand, realistic generations, i.e. samples from a region with high probability, do not change as much. The adversarial criterion for deterministic AAEs is difficult to optimise when the dimensionality of $Z$ is high. We observe that during training our AAEs and DAAEs, the empirical standard deviation of $\mathbf{z} \sim Q_{\phi}(Z|X)$ is less than 1, which means that $\hat{P}(Z)$ fails to approximate $P(Z)$ as closely as was achieved with the VAE and DVAE. However, this means that the effect of MCMC sampling is more pronounced, with the quality of all samples noticeably improving after a few steps. As a side-effect of the suboptimal solution learned by the networks, the denoising properties of the DAAE are more noticeable with the novel samples.}
  \label{all_samples}
\end{figure}

For evaluation, we generate novel samples from the decoder using
\(\mathbf{z}\) initially sampled from \(P(Z)\); we also show spherical
interpolations \citep{white2016sampling} between four images of the
testing split, as depicted in Figure \ref{vae_interp}. We then perform
several steps of MCMC sampling on the novel samples and interpolations.
During this process, we use the training mode of batch normalisation
\citep{ioffe2015batch}, i.e., we normalise the inputs using minibatch
rather than population statistics, as the normalisation can partially
compensate for poor initial inputs (see Figure \ref{all_samples}) that
are far from the training distribution. We compare novel samples between
all models below, and leave further interpolation results to Figures
\ref{interp_dvae} and \ref{interp_aae} of the supplementary material.

\subsection{Samples}\label{samples}

\section{Conclusion}\label{conclusion}

Autoencoders consist of a decoder, \(d(Z;\theta)\) and an encoder,
\(e(X;\phi)\) function, where \(\phi\) and \(\theta\) are learned
parameters. Functions \(e(X;\phi)\) and \(d(Z;\theta)\) may be used to
draw samples from the conditional distributions \(P_{\theta}(X|Z)\) and
\(Q_{\phi}(Z|X)\)
\citep{bengio2014deep, bengio2013generalized, rezende2014stochastic},
where \(X\) refers to the space of observed samples and \(Z\) refers to
the space of latent samples. The encoder distribution,
\(Q_{\phi}(Z|X)\), maps data samples from the data generating
distribution, \(P(X)\), to a latent distribution, \(\hat{P}(Z)\). The
decoder distribution, \(P_{\theta}(X|Z)\), maps samples from
\(\hat{P}(Z)\) to \(P(X)\). We are concerned with
\textit{generative autoencoders}, which we define to be a family of
autoencoders where regularisation is used during training to encourage
\(\hat{P}(Z)\) to be close to a known prior \(P(Z)\). Commonly it is
assumed that \(\hat{P}(Z)\) and \(P(Z)\) are similar, such that samples
from \(P(Z)\) may be used to sample a decoder \(P_{\theta}(X|Z)\); we do
not make the assumption that \(\hat{P}(Z)\) and \(P(Z)\) are
``sufficiently close'' \citep{rezende2014stochastic}. Instead, we derive
an MCMC process, whose stationary distribution is \(\hat{P}(Z)\),
allowing us to directly draw samples from \(\hat{P}(Z)\). By
conditioning on samples from \(\hat{P}(Z)\), samples drawn from
\(\mathbf{x} \sim P_{\theta}(X|Z)\) are more consistent with the
training data.

In our experiments, we compare samples
\(\mathbf{x} \sim P_{\theta}(X|Z=z_0)\), \(\mathbf{z}_0 \sim P(Z)\) to
\(\mathbf{x} \sim P_{\theta}(X|Z=z_i)\) for \(i=\{1,5,10\}\), where
\(\mathbf{z}_i\)'s are obtained through MCMC sampling, to show that MCMC
sampling improves initially poor samples (see Figure \ref{all_samples}).
We also show that artifacts in \(\mathbf{x}\) samples induced by
interpolations across the latent space can also be corrected by MCMC
sampling see (Figure \ref{vae_interp}). We further validate our work by
showing that the denoising properties of denoising generative
autoencoders are best revealed by the use of MCMC sampling.

Our MCMC sampling process is straightforward, and can be applied easily
to existing generative autoencoders. This technique is orthogonal to the
use of more powerful posteriors in AAEs \citep{makhzani2015adversarial}
and VAEs \citep{kingma2016improving}, and the combination of both could
result in further improvements in generative modeling. Finally, our
basic MCMC process opens the doors to apply a large existing body of
research on sampling methods to generative autoencoders.

\subsubsection*{Acknowledgements}
We would like to acknowledge the EPSRC for funding through a Doctoral
Training studentship and the support of the EPSRC CDT in
Neurotechnology.

\bibliography{../references.bib}
\bibliographystyle{iclr2017_conference}

\newpage
\raggedbottom
\appendix
\part*{Supplementary Material}
\section{\texorpdfstring{Proof that \(T(Z_{t+1}|Z_t) > 0\)
\label{proof}}{Proof that T(Z\_\{t+1\}\textbar{}Z\_t) \textgreater{} 0 }}\label{proof-that-tz_t1z_t-0}

\textbf{For \(\boldsymbol{P_\theta(X|Z) > 0}\) we require that all
possible \(\mathbf{x} \boldsymbol{\in X \subseteq \mathbb{R}^a}\) may be
generated by the network.} Assuming that the model \(P_\theta(X|Z)\) is
trained using a sufficient number of training samples,
\(\mathbf{x} \in X_{train} = X\), and that the model has infinite
capacity to model \(X_{train}=X\), then we should be able to draw any
sample \(\mathbf{x} \in X_{train}=X\) from \(P_\theta(X|Z)\). In reality
\(X_{train} \subseteq X\) and it is not possible to have a model with
infinite capacity. However, \(P_\theta(X|Z)\) is modeled using a deep
neural network, which we assume has sufficient capacity to capture the
training data well. Further, deep neural networks are able to
interpolate between samples in very high dimensional spaces
\citep{radford2015unsupervised}; we therefore further assume that if we
have a large number of training samples (as well as large model
capacity), that almost any \(\mathbf{x} \in X\) can be drawn from
\(P_\theta(X|Z)\).

Note that if we wish to generate human faces, we define \(X_{all}\) to
be the space of all possible faces, with distribution \(P(X_{all})\),
while \(X_{train}\) is the space of faces made up by the training data.
Then, practically even a well trained model which learns to interpolate
well only captures an \(X\), with distribution
\(\int P_\theta(X|Z) \hat{P}(Z) dZ\), where \(X_{train}\) \(\subseteq\)
\(X\) \(\subseteq\) \(X_{all}\), because \(X\) additionally contains
examples of interpolated versions of \(\mathbf{x} \sim P(X_{train})\).

\textbf{For \(\boldsymbol{Q_\phi(Z|X) > 0}\) it must be possible to
generate all possible
\(\mathbf{z} \boldsymbol{\in Z \subseteq \mathbb{R}^b}\).}
\(Q_\phi(Z|X)\) is described by the function
\(e(\cdot;\phi) : X \rightarrow Z\). To ensure that \(Q_\phi(Z|X) > 0\),
we want to show that the function \(e(X;\phi)\) allows us to represent
all samples of \(z \in Z\). VAEs and AAEs each construct \(e(X;\phi)\)
to produce \(z \in Z\) in different ways.

The output of the encoder of a VAE, \(e_{VAE}(X;\phi)\) is
\(\mathbf{z} = \boldsymbol{\mu} + \boldsymbol{\epsilon}\odot\boldsymbol{\sigma}\),
where
\(\boldsymbol{\epsilon} \sim \mathcal{N}(\mathbf{0}, \mathbf{I})\). The
output of a VAE is then always Gaussian, and hence there is no
limitation on the \(\mathbf{z}\)'s that \(e_{VAE}(X;\phi)\) can produce.
This ensures that \(Q_\phi(Z|X)>0\), provided that
\(\boldsymbol{\sigma} \neq \mathbf{0}\).

The encoder of our AAE, \(e_{AAE}(X;\phi)\), is a deep neural network
consisting of multiple convolutional and batch normalisation layers. The
final layer of the \(e_{AAE}(X;\phi)\) is a fully connected layer
without an activation function. The input to each of the \(M\) nodes in
the fully connected layer is a function \(f_{i=1...M}(\mathbf{x})\).
This means that \(\mathbf{z}\) is given by:
\(\mathbf{z} = \mathbf{a}_1f_1(\mathbf{x}) + \mathbf{a}_2f_2(\mathbf{x}) + ... + \mathbf{a}_Mf_M(\mathbf{x})\),
where \(\mathbf{a}_{i=1...M}\) are the learned weights of the fully
connected layer. We now consider three cases:

\textbf{Case 1:} If \(\mathbf{a}_i\) are a complete set of bases for
\(Z\) then it is possible to generate any \(\mathbf{z} \in Z\) from an
\(\mathbf{x} \in X\) with a one-to-one mapping, provided that
\(f_i(\mathbf{x})\) is not restricted in the values that it can take.

\textbf{Case 2:} If \(\mathbf{a}_i\) are an overcomplete set of bases
for \(Z\), then the same holds, provided that \(f_i(\mathbf{x})\) is not
restricted in the values that it can take.

\textbf{Case 3:} If \(\mathbf{a}_i\) are an undercomplete set of bases
for \(Z\) then it is not possible to generate all \(\mathbf{z} \in Z\)
from \(\mathbf{x} \in X\). Instead there is a many (X) to one (Z)
mapping.

For \(Q_\phi(Z|X) > 0\) our network must learn a complete or
overcomplete set of bases and \(f_i(x)\) must not be restricted in the
values that it can take \(\forall i\). The network is encouraged to
learn an overcomplete set of bases by learning a large number of
\(\mathbf{a}_i\)'s---specifically \(M = 8192\) when basing our network
on the DCGAN architecture \citep{radford2015unsupervised}---more that
\(40\) times the dimensionality of \(Z\). By using batch normalisation
layers throughout the network, we ensure that values of \(f_i(x)\) are
spread out, capturing a \textit{close-to-Gaussian} distribution
\citep{ioffe2015batch}, encouraging infinite support.

We have now shown that, under certain reasonable assumptions,
\(P_\theta(X|Z) > 0\) and \(Q_\phi(Z|X) > 0\), which means that
\(T(Z_{t+1}|Z_t) > 0\), and hence we can get from any \(Z\) to any
another \(Z\) in only one step. Therefore the Markov chain described by
the transition operator \(T(Z_{t+1}|Z_t)\) defined in Equation
(\ref{trans_operator}) is both irreducible and aperiodic, which are the
necessary conditions for ergodicity.

\section{CelebA}\label{celeba}

\subsection{Interpolations}\label{interpolations}

\begin{figure}[H]
  \centering
  \begin{subfigure}[]{0.97\linewidth}
    \includegraphics[width=\linewidth]{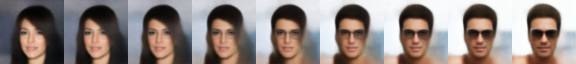}
    \subcaption{DVAE (initial)}
  \end{subfigure}
  \\[0.4ex]
  \begin{subfigure}[]{0.97\linewidth}
    \includegraphics[width=\linewidth]{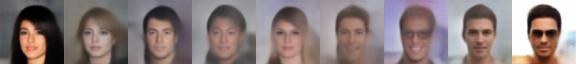}
    \subcaption{DVAE (5 steps)}
  \end{subfigure}
  \\[2ex]
  \begin{subfigure}[]{0.97\linewidth}
    \includegraphics[width=\linewidth]{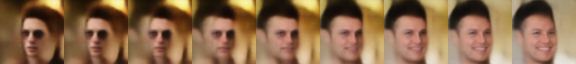}
    \subcaption{DVAE (initial)}
  \end{subfigure}
  \\[0.4ex]
  \begin{subfigure}[]{0.97\linewidth}
    \includegraphics[width=\linewidth]{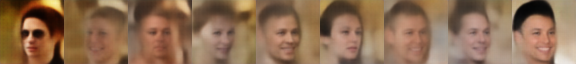}
    \subcaption{DVAE (5 steps)}
  \end{subfigure}
  \caption{Interpolating between two faces using (a-d) a DVAE. The top rows (a, c) for each face is the original interpolation, whilst the second rows (b, d) are the result of 5 steps of MCMC sampling applied to the latent samples that were used to generate the original interpolation. The only qualitative difference when compared to VAEs (see Figure \ref{all_samples}) is a desaturation of the generated images.}
  \label{interp_dvae}
\end{figure}

\begin{figure}
  \begin{subfigure}[]{0.97\linewidth}
    \includegraphics[width=\linewidth]{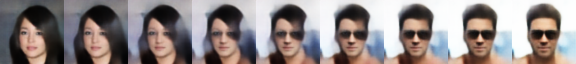}
    \subcaption{AAE (initial)}
  \end{subfigure}
  \\[0.4ex]
  \begin{subfigure}[]{0.97\linewidth}
    \includegraphics[width=\linewidth]{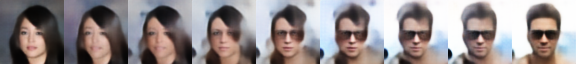}
    \subcaption{AAE (5 steps)}
  \end{subfigure}
  \\[2ex]
  \begin{subfigure}[]{0.97\linewidth}
    \includegraphics[width=\linewidth]{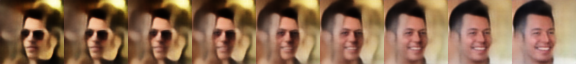}
    \subcaption{AAE (initial)}
  \end{subfigure}
  \\[0.4ex]
  \begin{subfigure}[]{0.97\linewidth}
    \includegraphics[width=\linewidth]{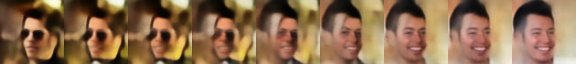}
    \subcaption{AAE (5 steps)}
  \end{subfigure}
  \\[2ex]
  \begin{subfigure}[]{0.97\linewidth}
    \includegraphics[width=\linewidth]{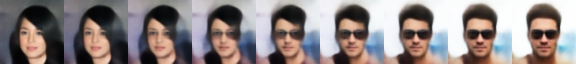}
    \subcaption{DAAE (initial)}
  \end{subfigure}
  \\[0.4ex]
  \begin{subfigure}[]{0.97\linewidth}
    \includegraphics[width=\linewidth]{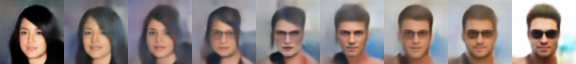}
    \subcaption{DAAE (5 steps)}
  \end{subfigure}
  \\[2ex]
  \begin{subfigure}[]{0.97\linewidth}
    \includegraphics[width=\linewidth]{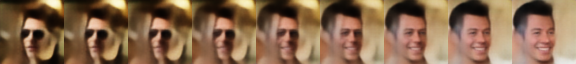}
    \subcaption{DAAE (initial)}
  \end{subfigure}
  \\[0.4ex]
  \begin{subfigure}[]{0.97\linewidth}
    \includegraphics[width=\linewidth]{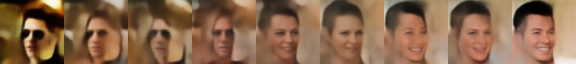}
    \subcaption{DAAE (5 steps)}
  \end{subfigure}
  \caption{Interpolating between two faces using (a-d) an AAE and (e-h) a DAAE. The top rows (a, c, e, g) for each face is the original interpolation, whilst the second rows (b, d, f, h) are the result of 5 steps of MCMC sampling applied to the latent samples that were used to generate the original interpolation. Although the AAE performs poorly (b, d), the regularisation effect of denoising can be clearly seen with the DAAE after applying MCMC sampling (f, h).}
  \label{interp_aae}
\end{figure}

\section{Street View House Numbers}\label{street-view-house-numbers}

\subsection{Samples}\label{samples}

\begin{figure}[H]
  \begin{subfigure}[]{0.21\linewidth}
    \includegraphics[width=\linewidth]{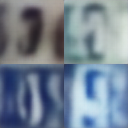}
    \subcaption{VAE (initial)}
  \end{subfigure}
  \hfill
  \begin{subfigure}[]{0.21\linewidth}
    \includegraphics[width=\linewidth]{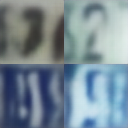}
    \subcaption{VAE (1 step)}
  \end{subfigure}
  \hfill
  \begin{subfigure}[]{0.21\linewidth}
    \includegraphics[width=\linewidth]{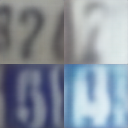}
    \subcaption{VAE (5 steps)}
  \end{subfigure}
  \hfill
  \begin{subfigure}[]{0.21\linewidth}
    \includegraphics[width=\linewidth]{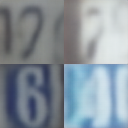}
    \subcaption{VAE (10 steps)}
  \end{subfigure}
  \\[2ex]
  \begin{subfigure}[]{0.21\linewidth}
    \includegraphics[width=\linewidth]{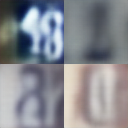}
    \subcaption{DVAE (initial)}
  \end{subfigure}
  \hfill
  \begin{subfigure}[]{0.21\linewidth}
    \includegraphics[width=\linewidth]{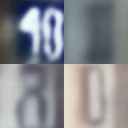}
    \subcaption{DVAE (1 step)}
  \end{subfigure}
  \hfill
  \begin{subfigure}[]{0.21\linewidth}
    \includegraphics[width=\linewidth]{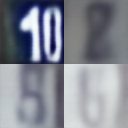}
    \subcaption{DVAE (5 steps)}
  \end{subfigure}
  \hfill
  \begin{subfigure}[]{0.21\linewidth}
    \includegraphics[width=\linewidth]{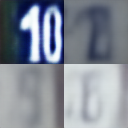}
    \subcaption{DVAE (10 steps)}
  \end{subfigure}
  \\[2ex]
  \begin{subfigure}[]{0.21\linewidth}
    \includegraphics[width=\linewidth]{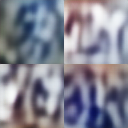}
    \subcaption{AAE (initial)}
  \end{subfigure}
  \hfill
  \begin{subfigure}[]{0.21\linewidth}
    \includegraphics[width=\linewidth]{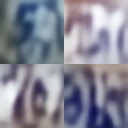}
    \subcaption{AAE (1 step)}
  \end{subfigure}
  \hfill
  \begin{subfigure}[]{0.21\linewidth}
    \includegraphics[width=\linewidth]{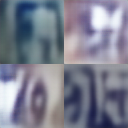}
    \subcaption{AAE (5 steps)}
  \end{subfigure}
  \hfill
  \begin{subfigure}[]{0.21\linewidth}
    \includegraphics[width=\linewidth]{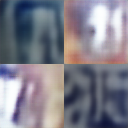}
    \subcaption{AAE (10 steps)}
  \end{subfigure}
  \\[2ex]
  \begin{subfigure}[]{0.21\linewidth}
    \includegraphics[width=\linewidth]{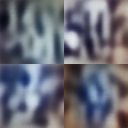}
    \subcaption{DAAE (initial)}
  \end{subfigure}
  \hfill
  \begin{subfigure}[]{0.21\linewidth}
    \includegraphics[width=\linewidth]{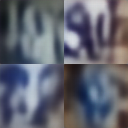}
    \subcaption{DAAE (1 step)}
  \end{subfigure}
  \hfill
  \begin{subfigure}[]{0.21\linewidth}
    \includegraphics[width=\linewidth]{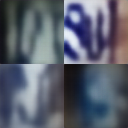}
    \subcaption{DAAE (5 steps)}
  \end{subfigure}
  \hfill
  \begin{subfigure}[]{0.21\linewidth}
    \includegraphics[width=\linewidth]{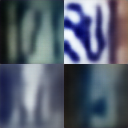}
    \subcaption{DAAE (10 steps)}
  \end{subfigure}
  \caption{Samples from a VAE (a-d), DVAE (e-h), AAE (i-l) and DAAE (m-p) trained on the SVHN dataset. The samples from the models imitate the blurriness present in the dataset. Although very few numbers are visible in the initial sample, the VAE and DVAE produce recognisable numbers from most of the initial samples after a few steps of MCMC sampling. Although the AAE and DAAE fail to produce recognisable numbers, the final samples are still a clear improvement over the initial samples.}
  \label{svhn_samples}
\end{figure}

\subsection{Interpolations}\label{interpolations-1}

\begin{figure}[H]
  \centering
  \begin{subfigure}[]{0.97\linewidth}
    \includegraphics[width=\linewidth]{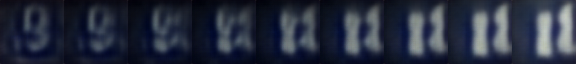}
    \subcaption{VAE (initial)}
  \end{subfigure}
  \\[0.4ex]
  \begin{subfigure}[]{0.97\linewidth}
    \includegraphics[width=\linewidth]{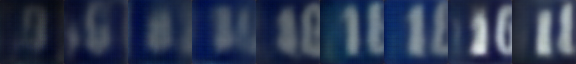}
    \subcaption{VAE (5 steps)}
  \end{subfigure}
  \\[2ex]
  \begin{subfigure}[]{0.97\linewidth}
    \includegraphics[width=\linewidth]{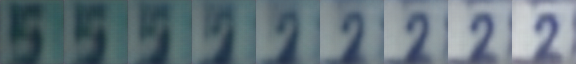}
    \subcaption{VAE (initial)}
  \end{subfigure}
  \\[0.4ex]
  \begin{subfigure}[]{0.97\linewidth}
    \includegraphics[width=\linewidth]{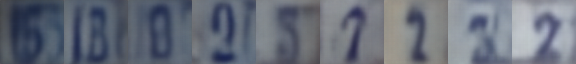}
    \subcaption{VAE (5 steps)}
  \end{subfigure}
  \\[2ex]
  \begin{subfigure}[]{0.97\linewidth}
    \includegraphics[width=\linewidth]{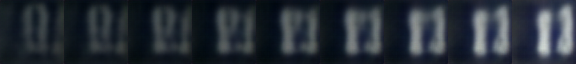}
    \subcaption{DVAE (initial)}
  \end{subfigure}
  \\[0.4ex]
  \begin{subfigure}[]{0.97\linewidth}
    \includegraphics[width=\linewidth]{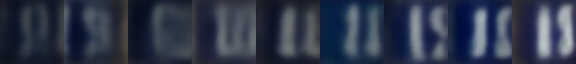}
    \subcaption{DVAE (5 steps)}
  \end{subfigure}
  \\[2ex]
  \begin{subfigure}[]{0.97\linewidth}
    \includegraphics[width=\linewidth]{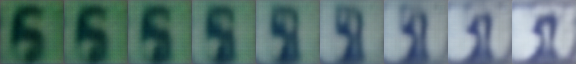}
    \subcaption{DVAE (initial)}
  \end{subfigure}
  \\[0.4ex]
  \begin{subfigure}[]{0.97\linewidth}
    \includegraphics[width=\linewidth]{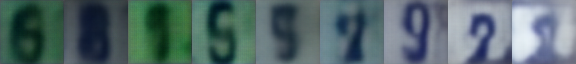}
    \subcaption{DVAE (5 steps)}
  \end{subfigure}
  \caption{Interpolating between Google Street View house numbers using (a-d) a VAE and (e-h) a DVAE. The top rows (a, c, e, g) for each house number are the original interpolations, whilst the second rows (b, d, f, h) are the result of 5 steps of MCMC sampling. If the original interpolation produces symbols that do not resemble numbers, as observed in (a) and (e), the models will attempt to move the samples towards more realistic numbers (b, f). Interpolation between 1- and 2-digit numbers in an image (c, g) results in a meaningless blur in the middle of the interpolation. After a few steps of MCMC sampling the models instead produce more recognisable 1- or 2-digit numbers (d, h). We note that when the contrast is poor, denoising models in particular can struggle to recover meaningful images (h).}
  \label{svhn_interp_vae}
\end{figure}

\end{document}